\documentclass[conference]{IEEEtran}
\IEEEoverridecommandlockouts

\usepackage{cite}
\usepackage{amsmath,amssymb,amsfonts}
\usepackage{algorithmic}
\usepackage{graphicx}
\usepackage{textcomp}
\usepackage{xcolor}
\def\BibTeX{{\rm B\kern-.05em{\sc i\kern-.025em b}\kern-.08em
    T\kern-.1667em\lower.7ex\hbox{E}\kern-.125emX}}

\usepackage{amssymb}
\usepackage{latexsym}
\usepackage{booktabs}
\usepackage{graphicx}
\usepackage{array}
\usepackage{cite}

\newcommand{\etal} {\textit{et~al.}}

\begin{document}

\title{Unsupervised Deep Learning for Handwritten Page Segmentation\\
}

\author{\IEEEauthorblockN{Ahmad Droby, Berat Kurar Barakat, Borak Madi, Reem Alaasam and Jihad El-Sana}

\IEEEauthorblockA{Ben-Gurion University of the Negev\\
             \{drobya,berat,borak,rym\}@post.bgu.ac.il\\
                    el-sana@cs.bgu.ac.il}}

\maketitle

\begin{abstract}
Segmenting handwritten document images into regions with homogeneous patterns is an important pre-processing step for many document images analysis tasks. Hand-labeling data to train a deep learning model for layout analysis requires significant human effort. In this paper, we present an unsupervised deep learning method for page segmentation, which revokes the need for annotated images. A siamese neural network is trained to differentiate between patches using their measurable properties such as number of foreground pixels, and average component height and width. The network is trained that spatially nearby patches are similar. The network's learned features are used for page segmentation, where patches are classified as main and side text based on the extracted features. We tested the method on a dataset of handwritten document images with quite complex layouts. Our experiments show that the proposed unsupervised method is as effective as typical supervised methods. 

\end{abstract}


\section{Introduction}
 Manually copying of manuscripts was the ultimate way scholars shared knowledge before the popularisation of the printing press. Notes were frequently added by scholars to the margin of pages, and often contribute valuable information concern the main text and the manuscript as a whole. In addition, the content of a manuscript's marginal notes help historians to analyze the authenticity, temporal, and geographical origin of the manuscript.

The increasing number of available digital scans of historical manuscripts, call for reliable automatic processing systems, which would allow historians and scholars to access and explore this knowledge more efficiently.

Page segmentation is an essential preprocessing step for many document image processing tasks. Due to the irregular structure, varying writing styles, and non-rectangular layout of historical handwritten documents~\cite{antonacopoulos2007special, likforman2007text}, segmenting them into main and side text poses a challenging research problem. 

Learning free based page layout analysis methods rely on human crafted features, such as connected component statistics~\cite{bukhari2012layout, bukhari2010document}, SIFT of points of interests~\cite{garz2011layout}, color and texture features~\cite{wei2013evaluation, chen2014page, wei2014hybrid}, etc. Due to the highly irregular structure and varying text style of historical handwritten documents, those methods do not generalize well. 
Therefore, researchers have been opting to learn those features instead. Page segmentation methods with a learning component generally outperform traditional learning free based methods. However, those methods require a large amount of manually annotated data for training in order to perform well. Obtaining such data is tedious and time-consuming; and in some cases requires domain experts.

We present an unsupervised deep learning method for page segmentation that utilizes measurable features such as spatial proximity, number of foreground pixels and average character height and width. The method first trains a siamese neural network model, $M$, then uses $M$ for feature extraction. A siamese network model contains two Convolutional Neural Networks (CNNs) with shared weights. The CNNs work in parallel on two different inputs to extract comparable feature vectors.
We train a siamese network to discriminate between patches with statistical differences of connected components; e.g., various number of foreground pixels and different background areas. Typically, in documents with side notes nearby patches belong to the same class (main or side text) with high probability. Based on this basic assumption, the network is trained that two spatially nearby patches are similar. Following training, we use the CNN component of the Siamese network to extract feature vector for every patch in a given page. 
The extracted feature vectors are then used to segment the page into main and side text regions. Our experimental results show that the accuracy of this method is comparable and in most cases surpasses the accuracy of supervised methods.

The rest of the paper is structured as follows. Section~\ref{sec:related_work} reviews related work. In Section~\ref{sec:method} we present our method in detail. Experimental results are reported in Section~\ref{sec:experiments}. Finally, conclusions are drawn, and future work is discussed in Section~\ref{sec:conclusion}.
l
]
\section{Related Work}
\label{sec:related_work}

Typically, page segmentation algorithms use features in order to segment pages into regions with homogeneous patterns. Existing page segmentation algorithms can be classified into two categories based on the type of used features.

\subsection{Hand-Crafted Features}
Traditional page segmentation algorithms rely upon hard-coded features, specification of documents structure, assumptions and statistical rules. Graz~\etal~\cite{garz2011layout} presented an approach to analyze the layout of handwritten documents using Scale Invariant Feature Transform (SIFT). The method uses Difference of Gaussian (DOG) to compute interest points, which guide the detection of layout entities. Finally, Support Vector Machine (SVM) is applied to classify the points into entity classes. Bukhari~\etal~\cite{bukhari2012layout} first extracts discriminative and simple features in the level of connected components, such as relative distance, foreground area, orientation, normalized height, and neighborhood information. Then a Multi-Layer Perceptron (MLP) classifies the connected components into side notes and main body texts. Finally, a voting scheme refines final classification results. Asi~\etal~\cite{abedAsi2014} proposed a learning-free approach for page segmentation of Arabic manuscripts. This is a two-step method: coarse segmentation and fine segmentation. Coarse segmentation utilizes Gabor texture filter and fine segmentation optimizes the results using energy minimization. Wong~\etal~\cite{Wong1982DocumentAS} use Run Length Smearing Algorithm (RLSA) for page segmentation. RLSA links together the neighboring areas that are black within predefined $c$ pixels. Two distinct bit maps are generated by applying RSLA row-by-row and column-by-column to a document. These maps are combined using 'AND' logical operator to produce segmented regions. These regions are then classified into text and non-text according to several criteria, such as black-white transitions and the total number of black pixels. Akiyama and Hagita \cite{AkiyamaHagita1990} divides an input document into smaller regions using basic features such as projection profiles, crossing counts, and circumscribed rectangles. These regions are then classified into headlines, text lines, and graphics regions. Apostolos~\cite{ApostolosAntonacopoulos1998} identifies background space surrounding the page regions and describes them using white tiles, which are horizontal rectangular white spaces. The algorithm can segment and identify regions with severe skew, but it does not classify them. Journet~\etal~\cite{Journet2008} extracts texture features and applies a multi-resolution analysis to avoid any assumption about the document's structure. Mehri~\etal~\cite{Mehri2013} segment a document into homogeneous regions by clustering texture features. Mehri~\etal~\cite{Mehri2013Comp} compared different approaches such as Gabor filters, auto-correlation function, and Grey Level Co-occurrence Matrix (GLCM). They conclude that for clustering and segmentation of document images, Gabor features are preformed best. Wei~\etal~\cite{wei2013evaluation} address segmentation as a pixel-level classification. Each pixel is a vector of its coordinates and its color values. They use SVM, MLP, and GMM to classify these vectors. Similarly, Chen~\etal~\cite{chen2014page} formulate layout analysis as a problem of pixel classification, where each pixel could belong to either periphery, background, text, or decoration. This method represents each pixel as a vector of its coordinates, color and texture. In addition, irrelevant features are removed by a feature selection algorithm
Chen~\etal~\cite{chen2014page} outperforms~\cite{wei2013evaluation} by including more features such as texture information and applying feature-selection algorithm for better classification result.

\subsection{Learned Features}
In the past decade, learning features using CNN has become the dominant approach in the page-layout analysis domain.

Chen~\etal~\cite{chen2015AutoEnc} apply convolutional autoencoders for learning the features from pixels. These features are used to train an SVM for page segmentation.
The classifier assigns to each pixel one of four classes: periphery, background, text block, and decoration. Later they applied SVM to classify superpixels instead of pixels ~\cite{chen2016AutoEncSuper} to reduce the classification time complexity. In addition, segmentation results are further refined in~\cite{chen2016AutoEncCRF} using Conditional Random Field (CRF) that utilizes local and contextual information. These works~\cite{chen2015AutoEnc,chen2016AutoEncCRF} consider feature extraction and classifying as two separate steps. On the other hand, \cite{chen2017} introduced an end-to-end method by combining feature learning and classifier training into one step.




Recently, Kurar~\etal~\cite{barakat2018binarization} and Alaasam~\etal~\cite{alaasam2019layout} trained Fully Convolutional Network (FCN) and siamese neural network, respectively, to apply page segmentation. Both Kurar~\etal~\cite{barakat2018binarization} and Alaasam~\etal~\cite{alaasam2019layout} reported their results on the same dataset that we use for evaluation.

\section{Method}
\label{sec:method}

\begin{figure}
    \centering
    \includegraphics[width=0.4\textwidth]{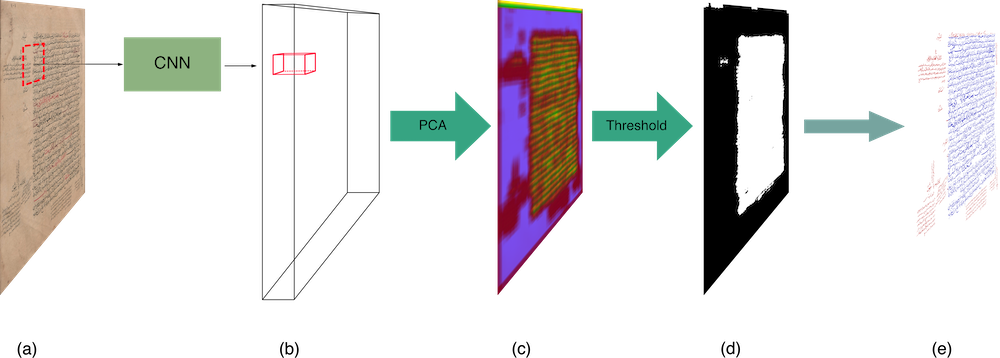}
    \caption{Method flow: (a) Input page, (b) the resulting feature map by applying the trained CNN on the input page image using a sliding window, (c) a visualization of the first three principal components of the feature map, (d) applying a threshold on the first and second principal components of the feature map to extract the main-text mask, and (e) the final segmentation of the page, where foreground pixels inside the main-text mask are determined to be part of the main-text; otherwise, they are part of the side text}
    \label{fig:method_flow}
\end{figure}

Our method is composed of two main steps: feature extraction and segmentation. Feature extraction is a crucial step in any layout analysis algorithm. We delegate this step to a CNN trained as a branch of a siamese network, which is then used to extract features from patches in a given page. The siamese network is trained using patches prepared according to multiple strategies. We apply principal component analysis to the feature map and use the first and the second principle components to guide classifying the map into two categories: main text and side notes.


\subsection{Data preparation}
Data preparation consists of generating patches of the size $200\times200$ pixels, cropped randomly from document images and labeling. Every pair of patches are labeled either similar or different based on a set of principles we discuss below. Patch size is estimated as four times the average character height in the input document images. 
Without loss of generality and by analogy with distance, we label similar pairs of patches by zero and different pairs by one. We use four strategies to generate pairs of image patches with labels. One of them is for similar pairs of patches and the remaining three are for different pairs. 
The principles used to generate and label the patches are dataset independent and generalize to other datasets with heterogeneous text line-heights.

Next we discuss the four strategies to generate pairs of image patches with their labels.

\subsubsection{Patches similar by spatial proximity}
\begin{figure}[h]
\centering
\includegraphics[width=4cm]{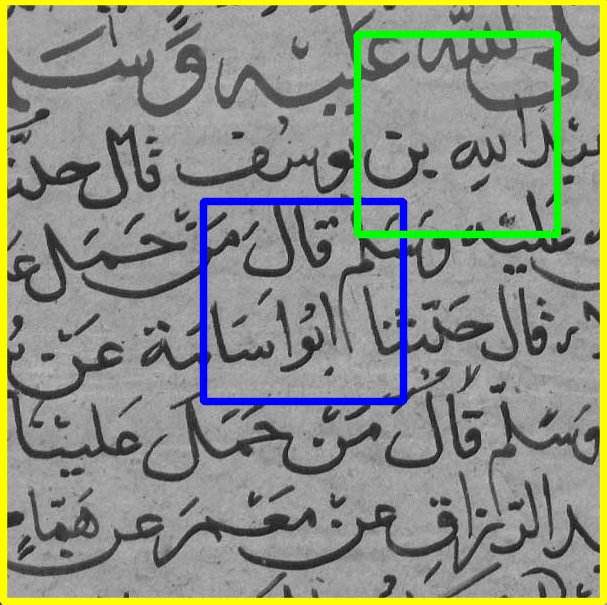}
\caption{Generating patches from document image}
\label{pssn}
\end{figure}

Patches are labeled by a simple principle, neighbouring patches are similar~\cite{danon2019unsupervised}. Given a document image we randomly sample a first patch, $p_1$ and an arbitrarily second patch, $p_2$, from the eight possible neighbouring locations around $p_1$ (see illustration in \figurename~\ref{pssn}). In order to avoid trivial solutions, we perturb the location of $p_2$ by a quarter of the patch's height. Naturally, some neighbouring patches are not similar (e.g. patches located between main and side text). However, such patches are rare enough relative to similar neighbouring patches to be considered as noise.

\subsubsection{Patches different by average component sizes}
Given randomly cropped two image patches, let $h_i$ and $w_i$ be the average component height and width of patch $i$, respectively, where $i\in \{1,2\}$. Our algorithm iteratively sample, at random, pairs of patches until the similarity score $s_1$ satisfies the following condition:
\begin{equation}
\label{similar_patches}
s_1=\frac{\min(h_1\times w_1,h_2\times w_2)}{\max(h_1\times w_1,h_2\times w_2)} < 0.5
\end{equation}

\begin{figure}[h]
\centering
\includegraphics[width=6cm]{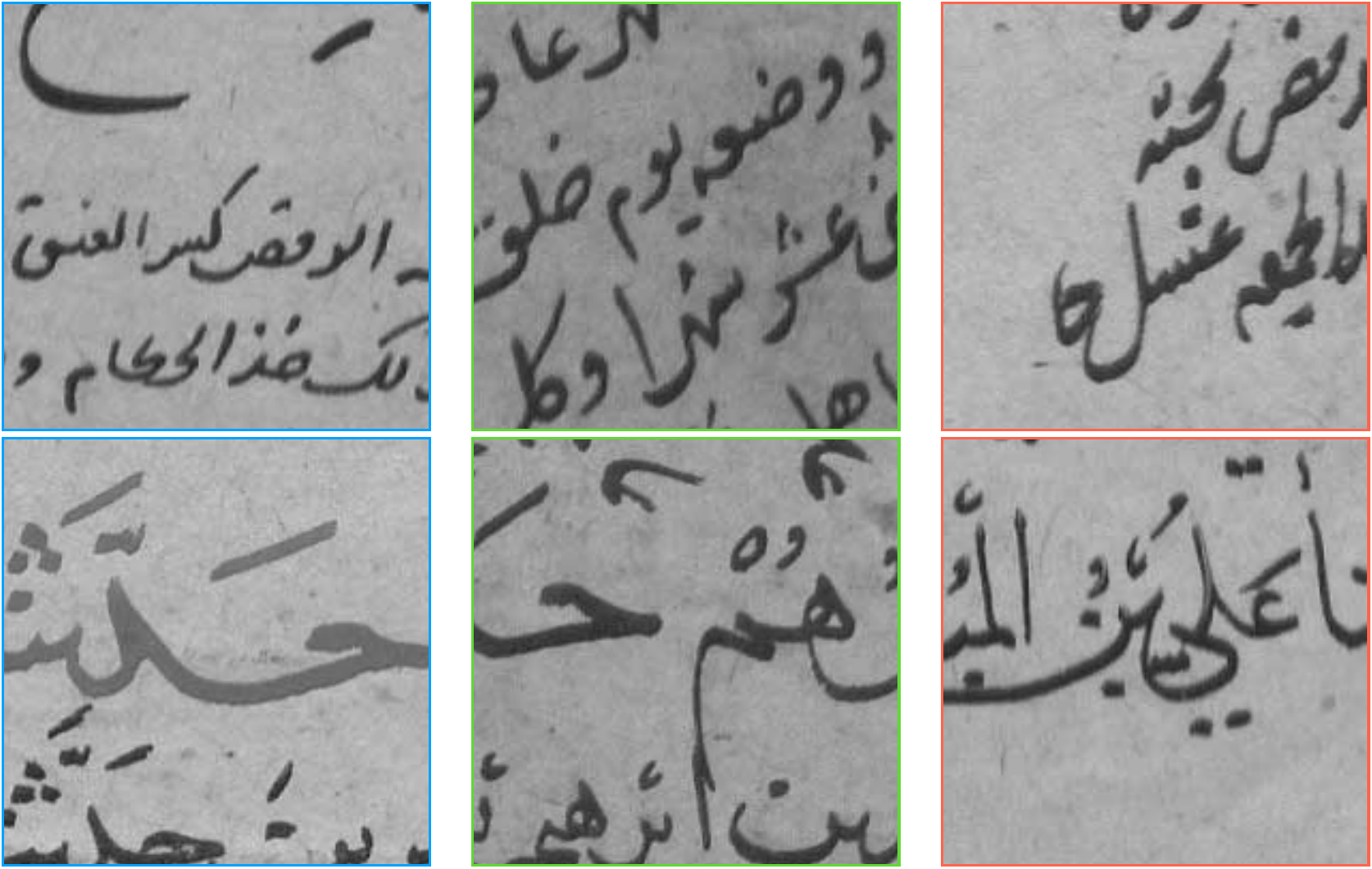}
\caption{Every column shows a pair of patches different by average component height and width. Such pairs train the machine to discriminate between the side text (above patches) and the main text areas (below patches).}
\label{pdbcs}
\end{figure}

In a loosely manner this strategy generates pairs of patches where one from main text area and the other from side text area (\figurename~\ref{pdbcs}). This is based on the assumption that the side text is written in a relatively small and restricted margins on the page, resulting in text with smaller font size. Therefore, the average component's height and width in side text area are relatively less than the average component's height and width in main text area. 

\subsubsection{Patches different by number of foreground pixels}
Due to the font size difference between main and side text, the number of foreground pixels in side text area is relatively less than the number of foreground pixels in main text area. This assumption is used in this strategy to differentiate between patches from main text area and  patches from side text areas.

Given randomly cropped two image patches, let $a_i$ be the number of foreground pixels in patch $i$, where $i\in \{1,2\}$. The algorithm continues selecting two random patches until the similarity score $s_2$ satisfies the following condition:
\begin{equation}
\label{different_patches}
s_2=\frac{\min(a_1,a_2)}{\max(a_1,a_2)}< 0.5
\end{equation}

\begin{figure}[h]
\centering
\includegraphics[width=6cm]{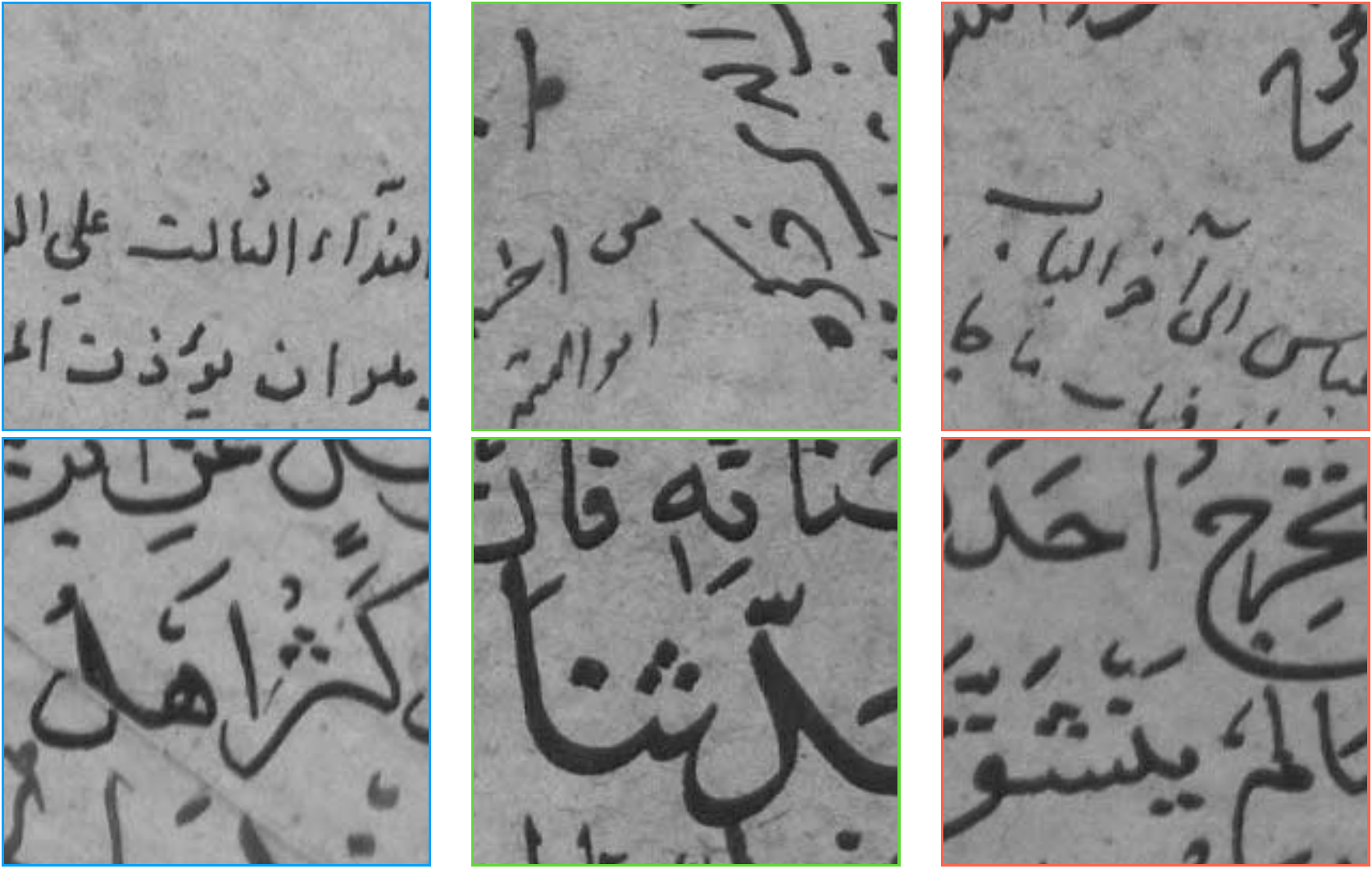}
\caption{Every column shows a pair of patches different by the number of foreground pixels. Such pairs train the machine to discriminate among the side text (above patches) and the main text areas (below patches).}
\label{pdbfa}
\end{figure}

In a loosely manner this strategy generates pairs of patches where one from main text area and the other from side text area, as illustrated in \figurename~\ref{pdbfa}. 

\subsubsection{Patches different by background area}
\begin{figure}[h]
\centering
\includegraphics[width=6cm]{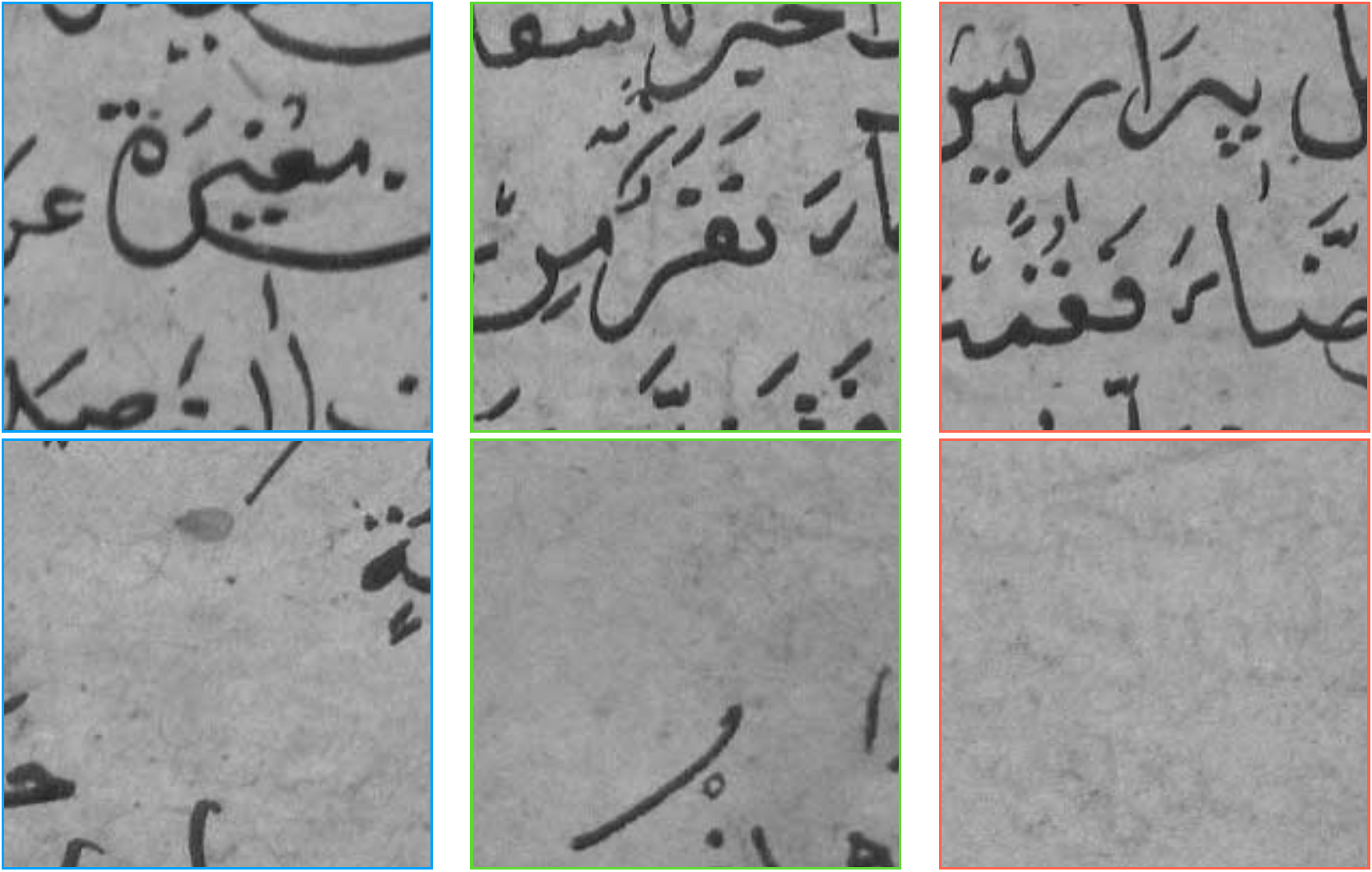}
\caption{Every column shows a pair of different patches. In a loosely manner, either of patches in each pair contain background area or foreground area. Such pairs train the machine to discriminate the text areas from the background areas.}
\label{pdbb}
\end{figure}

A significant difference between background areas and text areas exists often in document images. This strategy iteratively sample pair of patches at random until one of the patches is from background area and the other from text area, as shown in Figure~\figurename~\ref{pdbb}. We assume a patch belongs to a background area if more than its half belongs to a background area.

\subsection{Siamese network}
We train a Siamese network with two identical CNN branches. The input is a pair of image patches of size $200 \times 200$. The CNN branches extract representations of the input patches. Subsequently, these representations are concatenated and fed into fully connected layers in order to classify whether the two image patches are similar or different (further details are given in Section~\ref{sec:experiments}.)

\subsection{Feature extraction}
We use one of the branches of the trained siamese network for feature extraction. 
This branch takes a patch of size $200 \times 200$ and applies a number of convolutional layers followed by two fully connected layers and outputs a feature vector of size $512$, as shown in Figure~\figurename~\ref{fig:siamese_arch}.

In the feature extraction step, a sliding window is used to extract a feature vector for each pixel in the input image using the CNN branch of the trained siamese network. As can be seen in \figurename~\ref{fig:method_flow}, the feature extraction step outputs a feature map of size $w \times h \times 512$, where $w$ and $h$ are the width and height of the input image respectively.

\subsection{Segmentation}
The obtained feature map is used to guide the segmentation of the page into main and side text regions. The construction of the feature map aims at representing the two segments differently to simplify the segmentation procedure.

We have investigated applying PCA on the feature map and study (visualize and analyze) the resulting subspace. The first and the second principal components lead to a good indication of main text location, where the values of the first and the second principal components are higher for main text areas than side text areas. Therefore, we thresholded the feature map based on the first and second principal components to segment the main-text; i.e., a pixel $p$ in the image is denoted main-text if the following condition holds:

\begin{align*}
PC_1(p) < T_1 \text{ and } PC_2(p) < T_2
\end{align*}
where $PC_i(p)$ is the $i$'th principal component of the feature vector at pixel $p$ and $T_1, T_2$ are predefined thresholds based on experimental results.


The network learns to extract meaningful information about the patches, such as text orientation, number background and foreground pixels, and connected component statistics. As a result, it extracts similar features from main text patches which are different from those extracted from side text patches, and similar features from side text patches which are different from those extracted from main text patches. We searched for a scheme to reduce the dimensions of the feature space to two while maintaining the distances between the data points. Since PCA does this well, we adopted it for dimension reduction. We have found that the first two components provide good results and the segmentation (int main and side text) is carried out using a simple threshold.




\section{Experiments}
\label{sec:experiments}
In this section we present the dataset we adopted for training and test, discuss training procedure, and analyse the obtained results.

\subsection{Dataset}
We have choose to evaluate our approach using the dataset presented by Bukhari~\etal~\cite{bukhari2012layout}. The dataset consists of 38 handwritten document images from 7 different historical Arabic Books. It is split as follows: 28 documents for training and the remaining 10 images are used for testing. The main-text and the side-text are labeled in each document in the dataset. To train the Siamese network we use 24 documents from the training set and the remaining 6 documents are used for validation.

\subsection{Training}
\begin{figure}[h]
\centering
\includegraphics[width=0.40\textwidth]{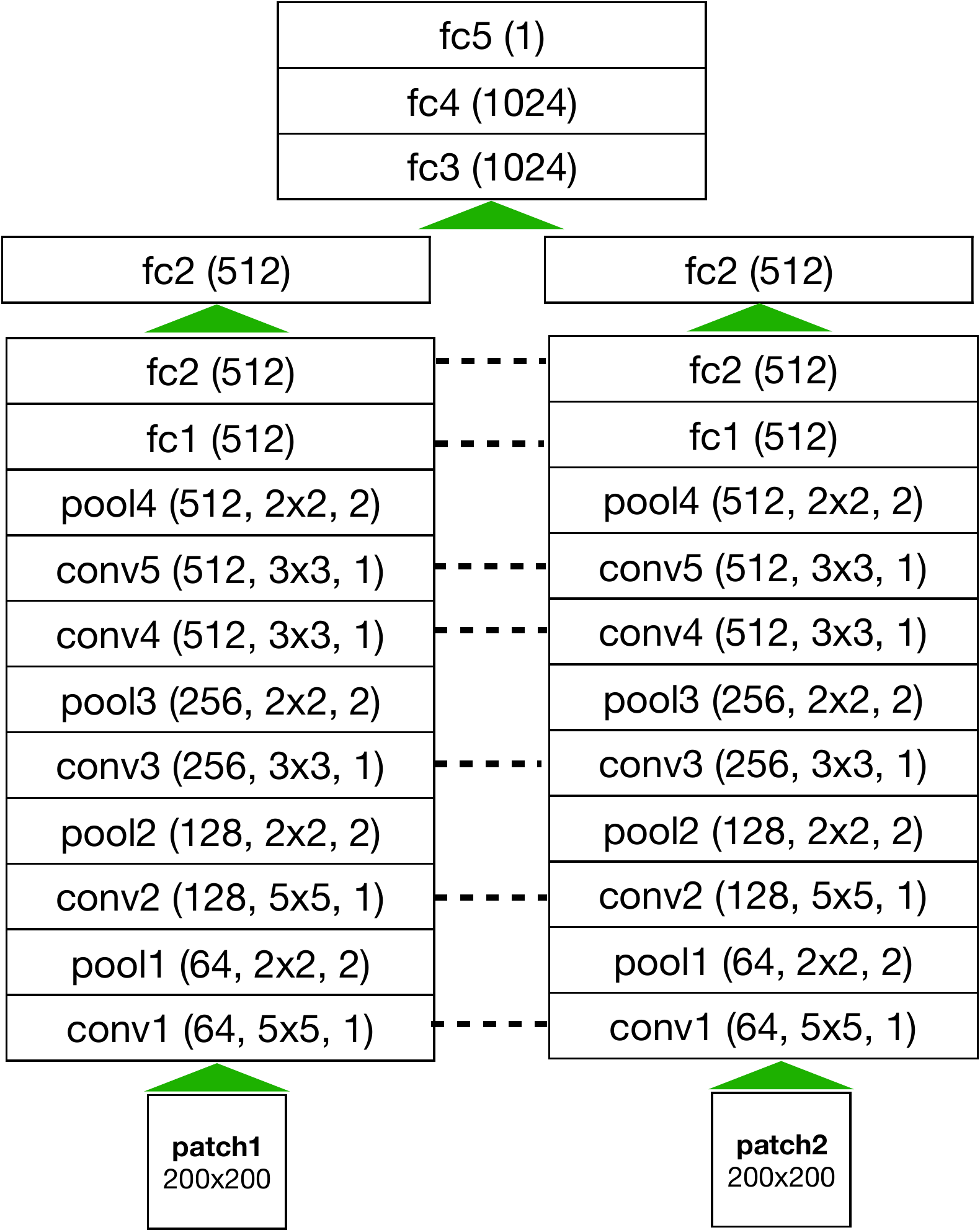}
\caption{Siamese architecture for classifying pairs as similar or different. Dotted lines stand for identical weights, conv stands for convolutional layer, fc stands for fully connected layer and pool is a max pooling layer.}
\label{fig:siamese_arch}
\end{figure}

We built the Siamese network's branches similar to the Alexnet~\cite{krizhevsky2012imagenet} model and through experiments we tune the hyperparameters to fit our problem. The final architecture consists of two CNN branches, each one has five convolutional layers as shown in \figurename~\ref{fig:siamese_arch}. Dotted lines indicate identical weights, and the numbers in parentheses represent the number of filters, filter size, and stride. All convolutional and fully connected layers are followed by ReLU activation functions, except fc5, which feeds into a sigmoid binary classifier. The learning rate is $0.00001$ and the optimizing algorithm is ADAM.

\begin{figure}[h]
\centering
\includegraphics[width=7.5cm]{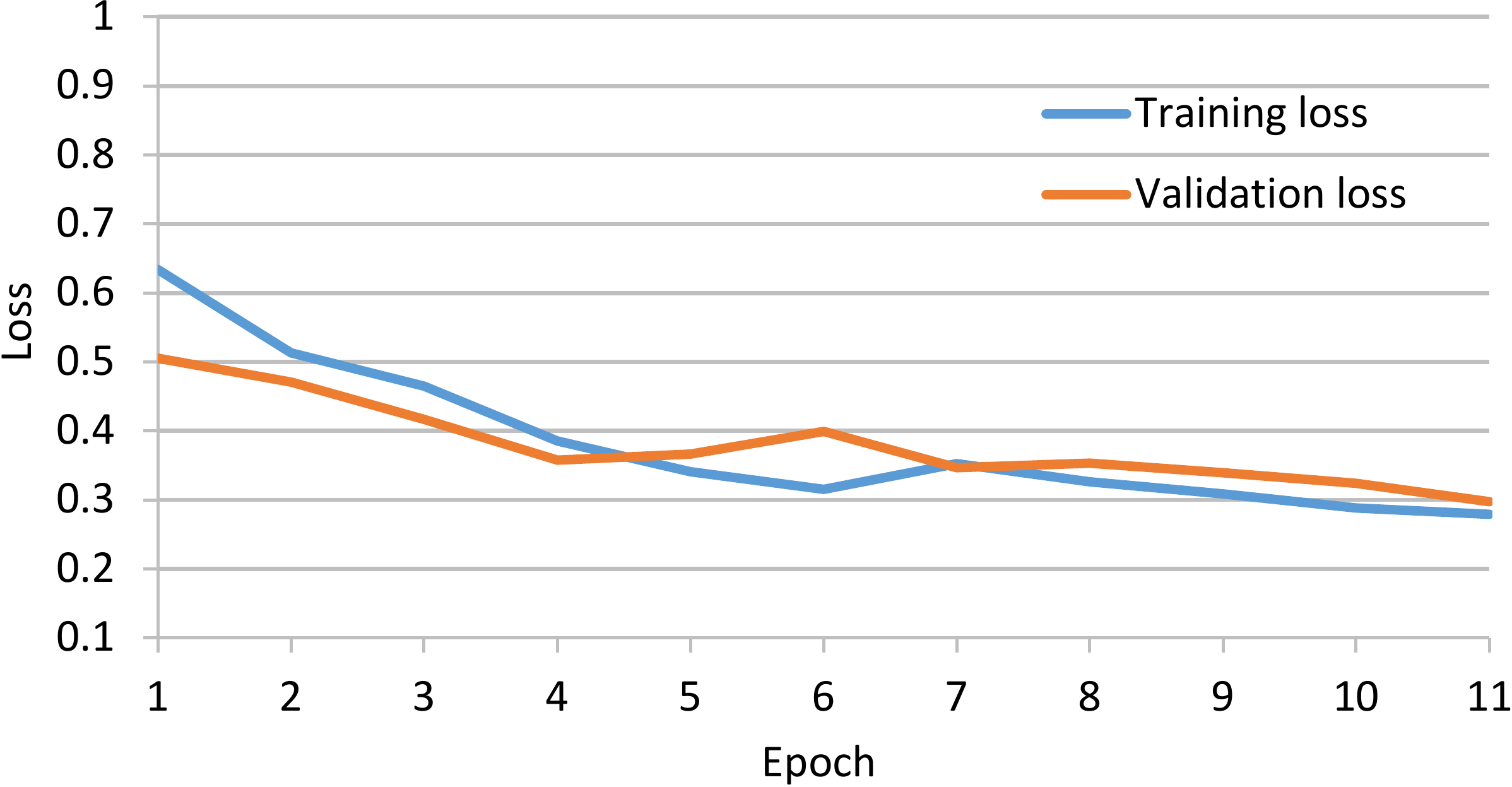}
\caption{Loss over the epochs of model training.}
\label{fig:loss}
\end{figure}

We trained this model from scratch using $60,000$ pairs with balanced classes and reached a validation loss value of $0.30$ after 11 epochs (\figurename~\ref{fig:loss}). When the training is done, we cut out a branch of the Siamese network to be used for feature extraction.

\subsection{Results}
We applied our method to segment the pages in the test set of the dataset into main and side text regions. The non-binarized images were used in the feature extraction step of the method. Extracting the feature vector for every possible patch in the image is expensive time-wise. Therefore, we used a sliding window with a step size of 50 pixels resulting in a feature map with dimensions smaller than the original image. In order to match the original image dimension, the feature map was resized with bi-linear interpolation.

In Table~\ref{tab:results} we compare the performance of the proposed method using F-measure against the layout analysis methods~\cite{bukhari2012layout, barakat2018binarization, alaasam2019layout}. Note that those three methods uses labeled data to train a ML model, while the proposed method is trained in an unsupervised manner. Our method outperformed both Bukhari~\etal~\cite{bukhari2012layout} and Kurar~\etal~\cite{barakat2018binarization} on both the main-text and the side-text. While it outperformed Alaasam~\etal~\cite{alaasam2019layout} on the side-text, we obtained slightly lower results on the main-text. However, it worth noting that Alaasam~\etal preformed post-processing on their results whereas we do not.

\begin{figure}
    \centering
    \begin{tabular}{>{\centering\arraybackslash}c m{0.1\textwidth} m{0.1\textwidth} m{0.1\textwidth}}
        {Input} & \includegraphics[width=0.1\textwidth]{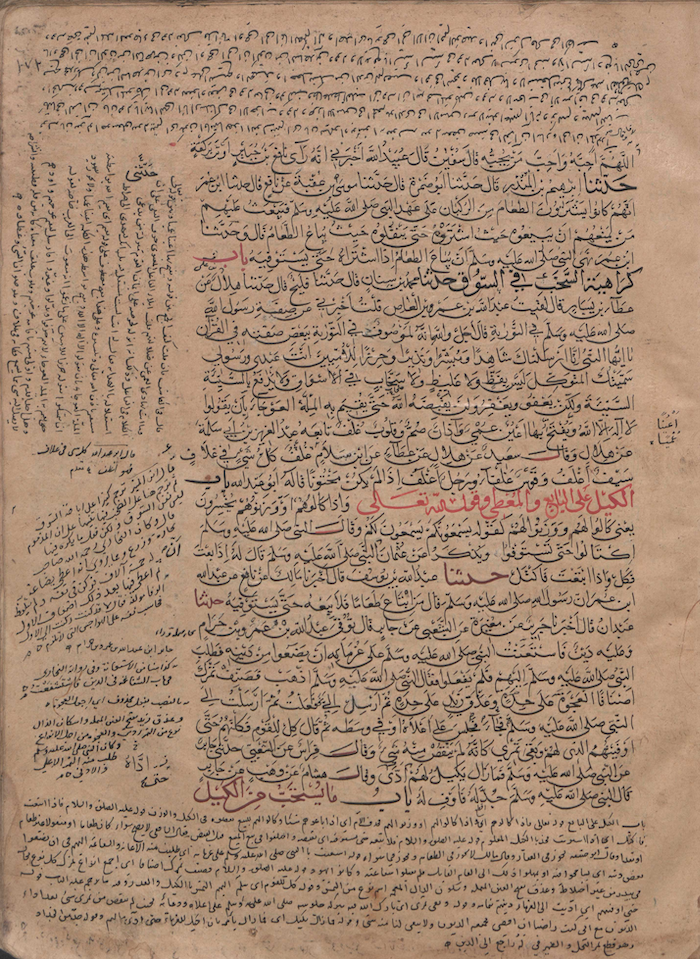} &
        \includegraphics[width=0.1\textwidth]{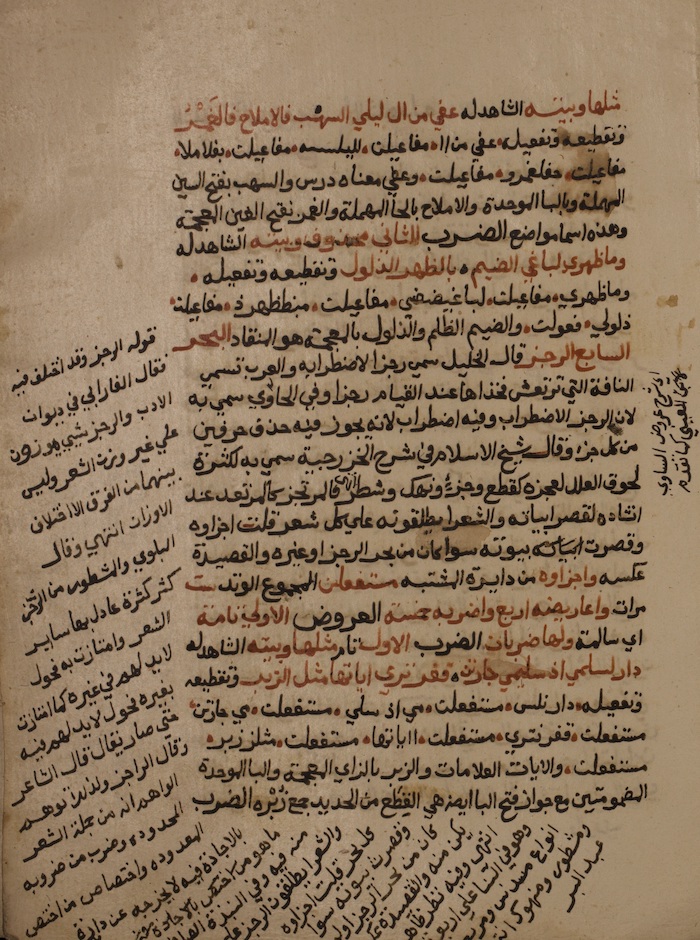} &
        \includegraphics[width=0.1\textwidth]{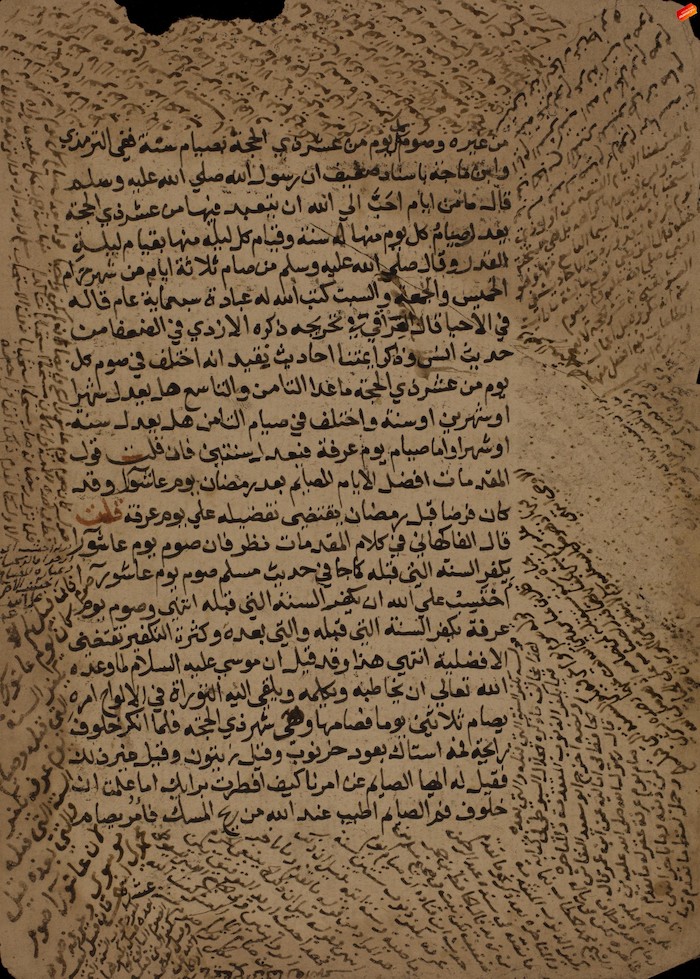}\\
        {Feature map} & \includegraphics[width=0.1\textwidth]{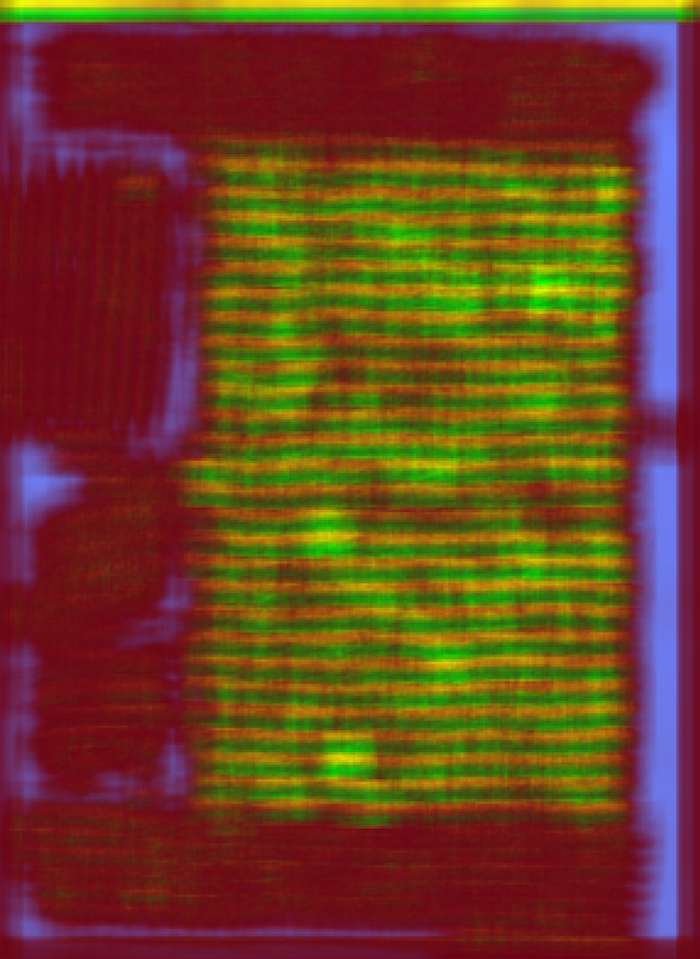} &
        \includegraphics[width=0.1\textwidth]{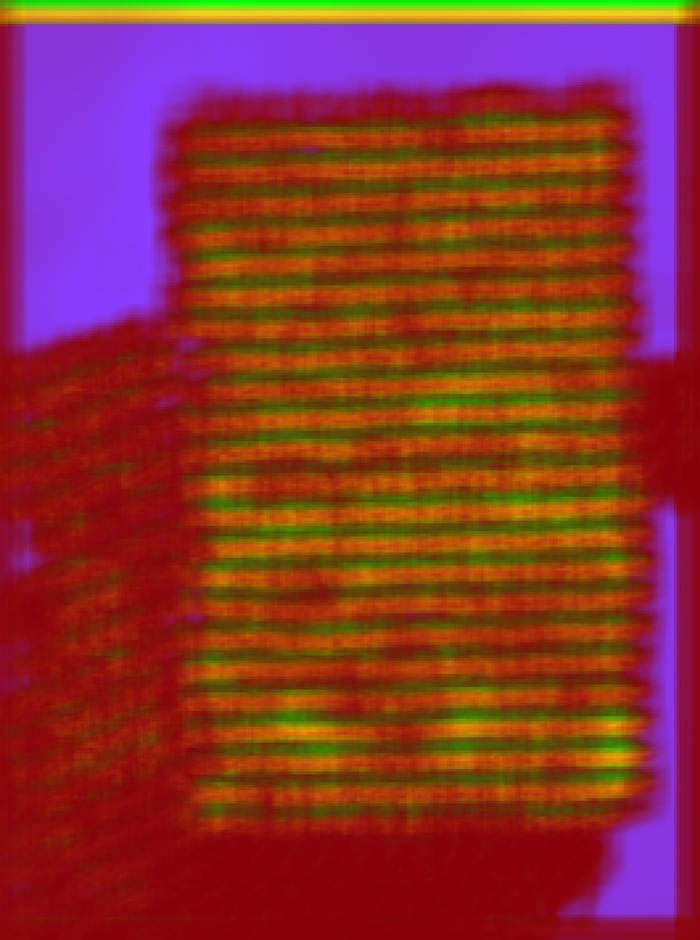} &
        \includegraphics[width=0.1\textwidth]{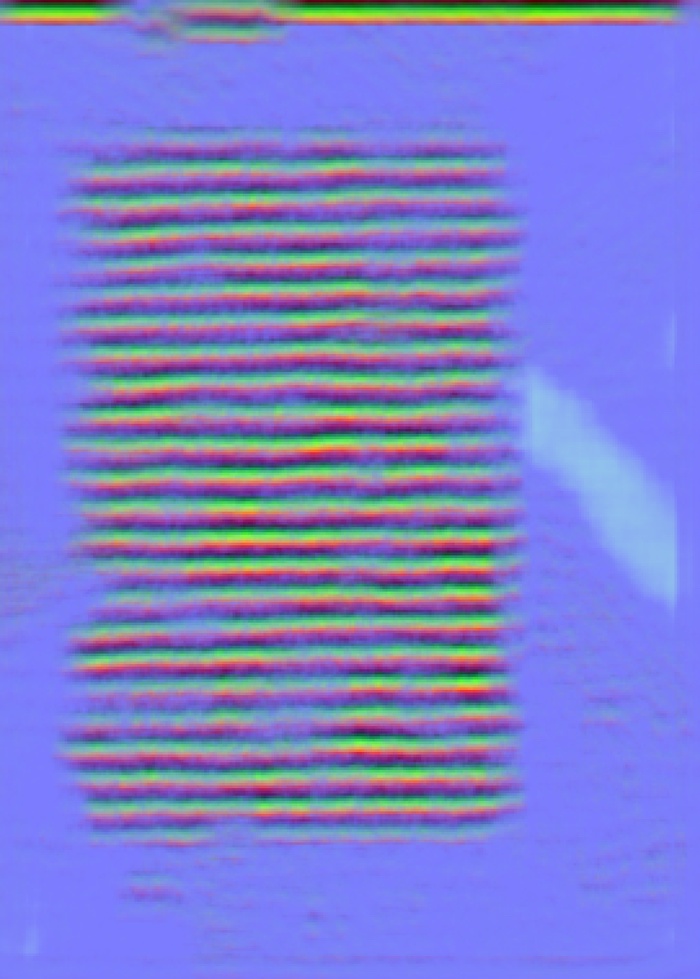}\\
        {Main-text} & \includegraphics[width=0.1\textwidth]{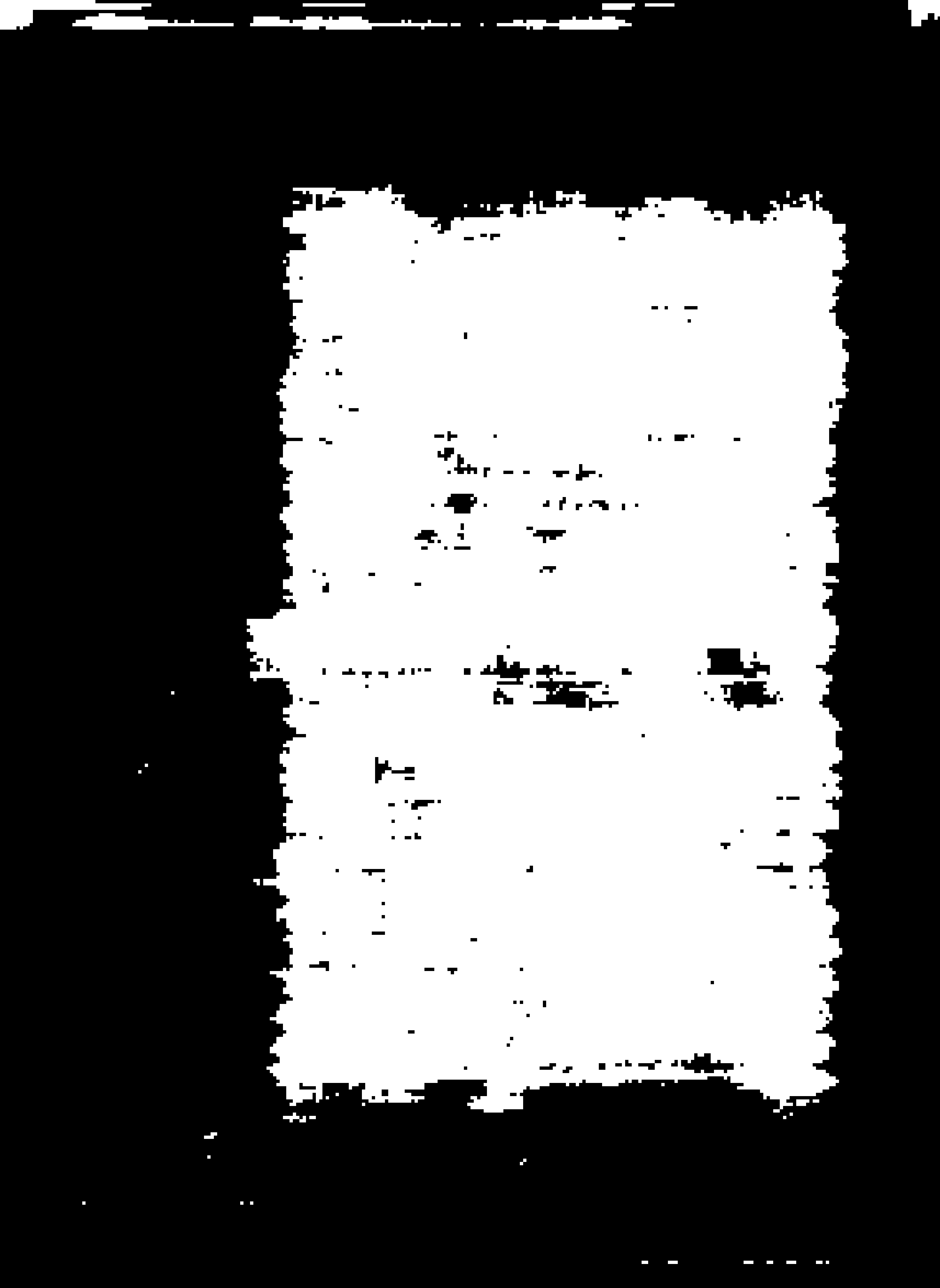} &
        \includegraphics[width=0.1\textwidth]{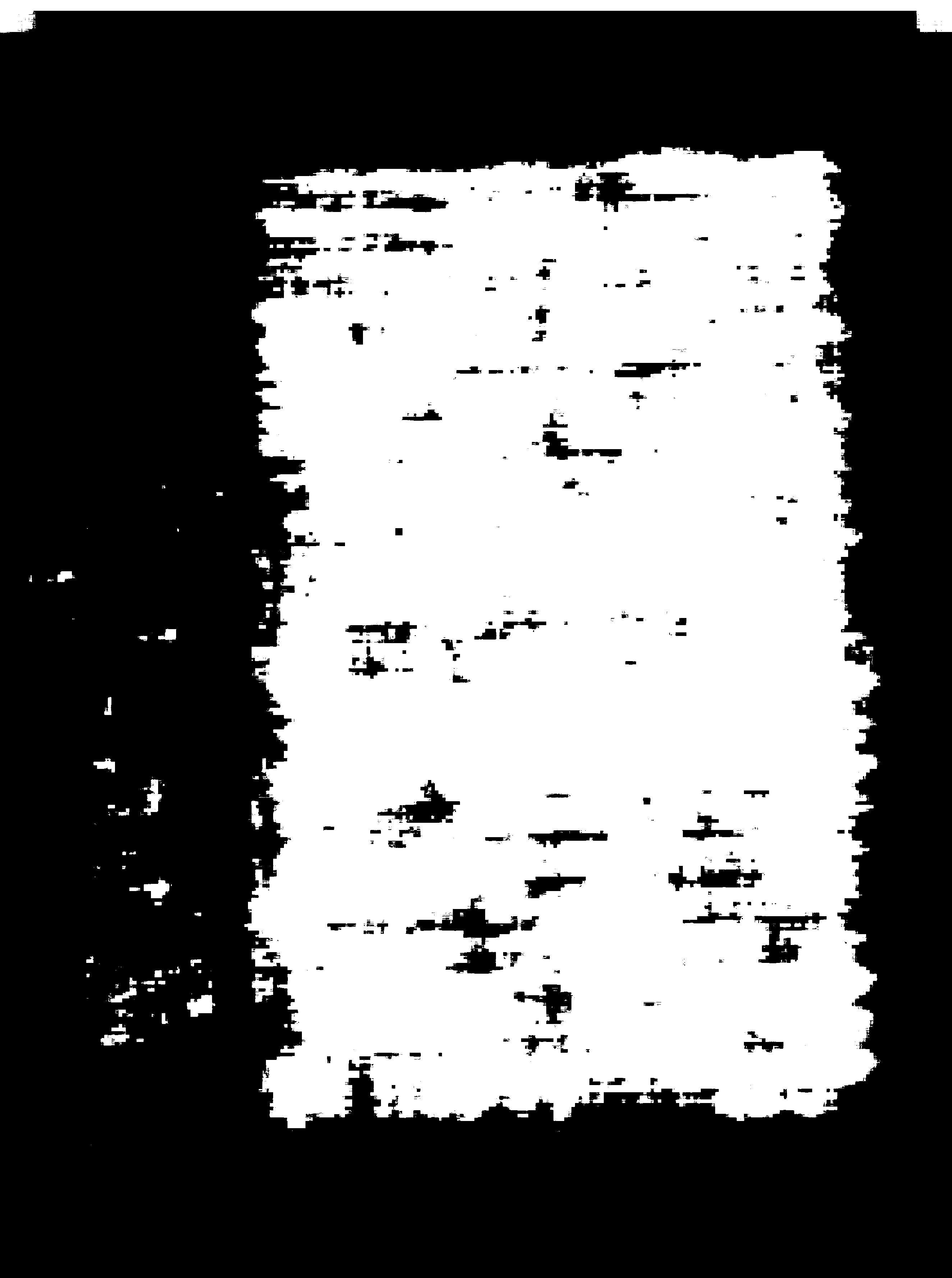} &
        \includegraphics[width=0.1\textwidth]{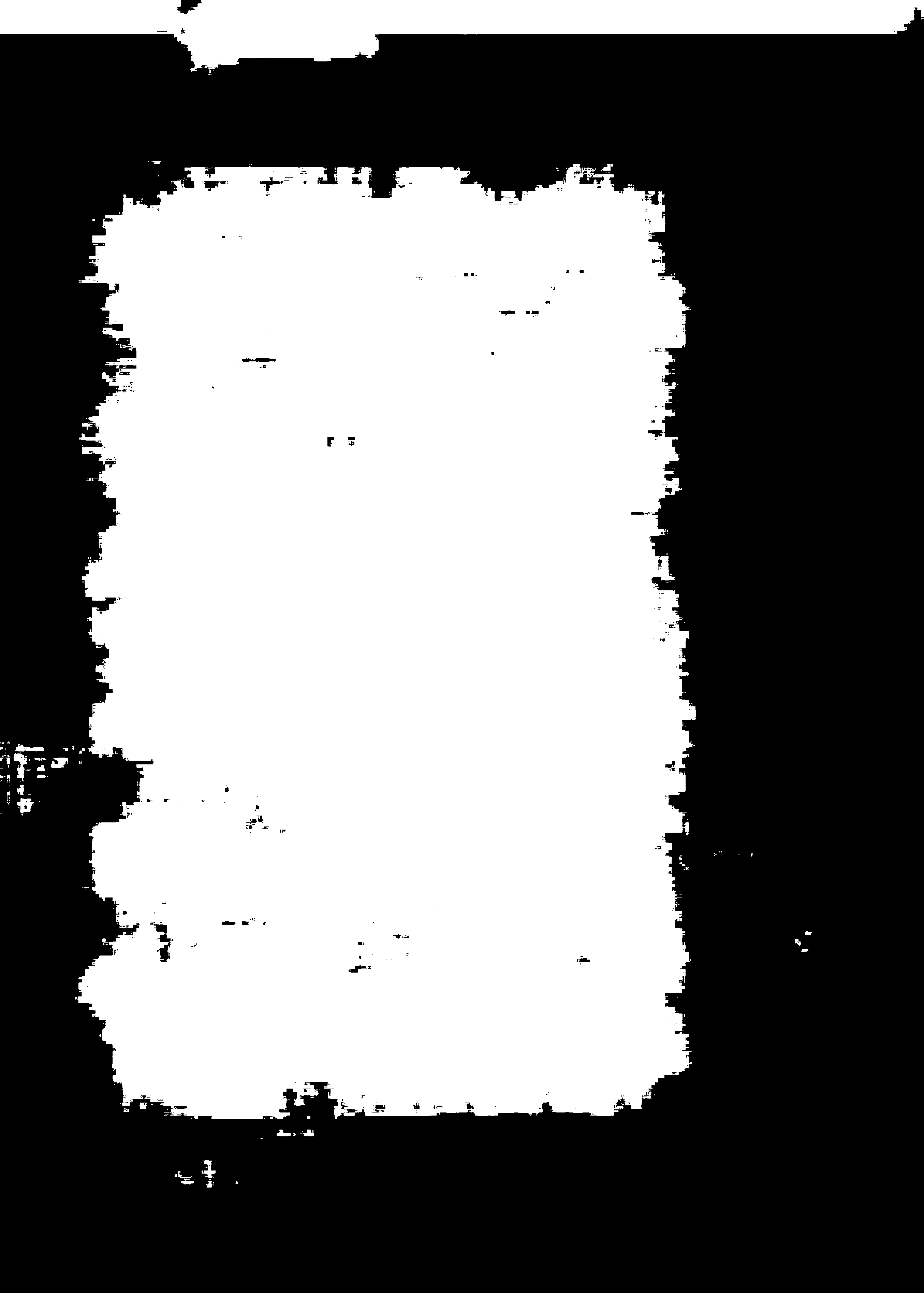}\\
        {Result} & \includegraphics[width=0.1\textwidth]{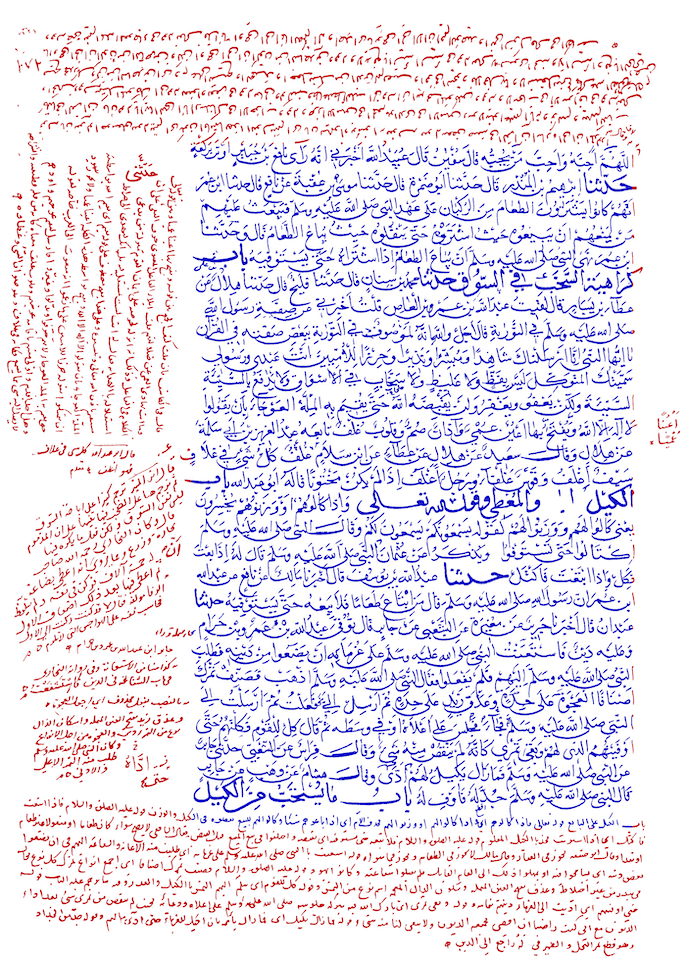} &
        \includegraphics[width=0.1\textwidth]{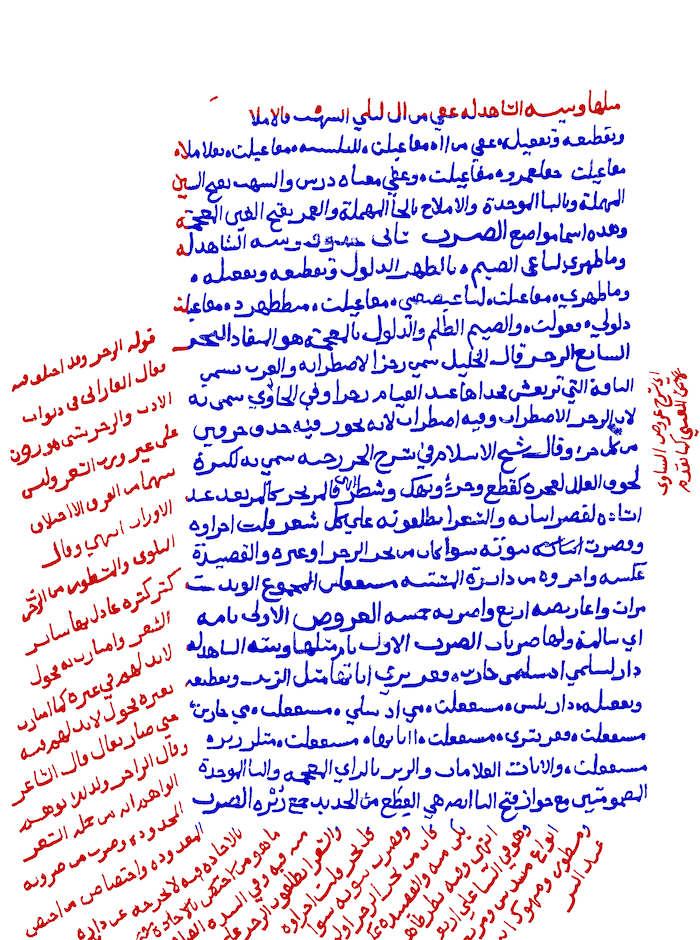} &
        \includegraphics[width=0.1\textwidth]{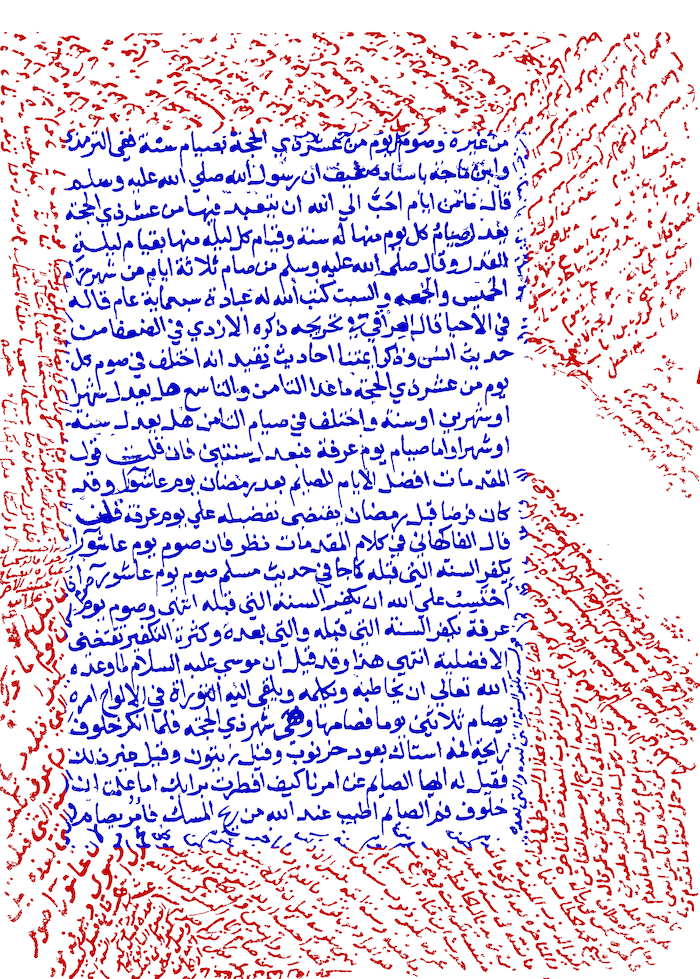}\\
        {Groundtruth} & \includegraphics[width=0.1\textwidth]{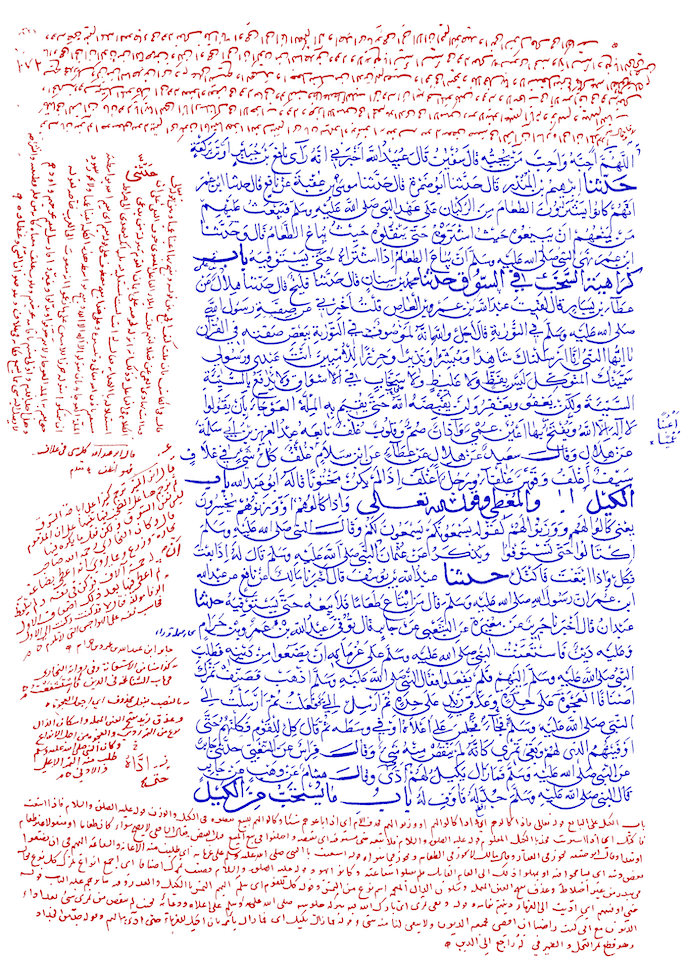} &
        \includegraphics[width=0.1\textwidth]{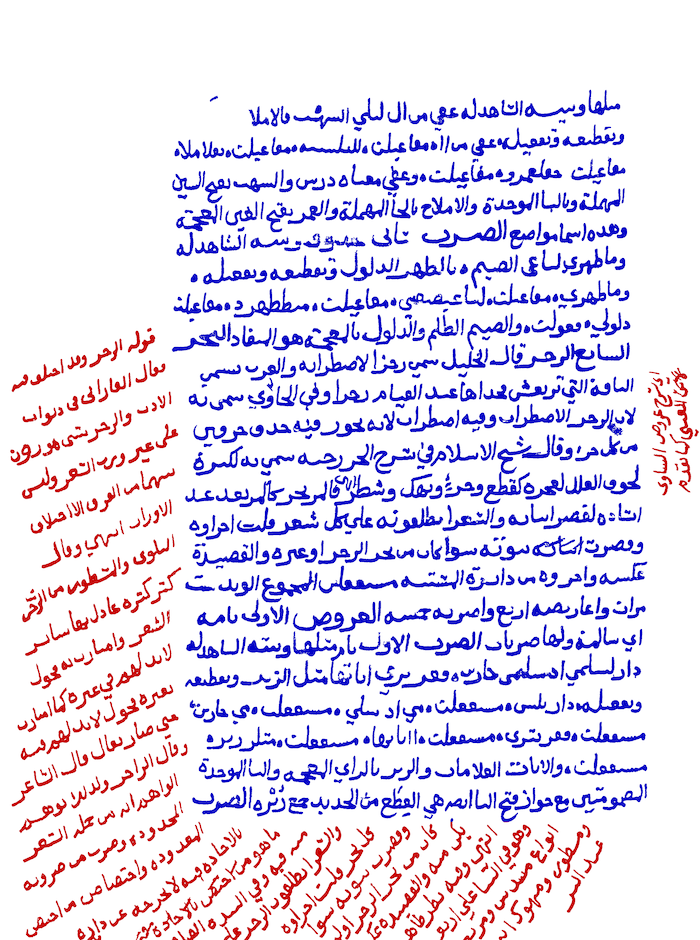} &
        \includegraphics[width=0.1\textwidth]{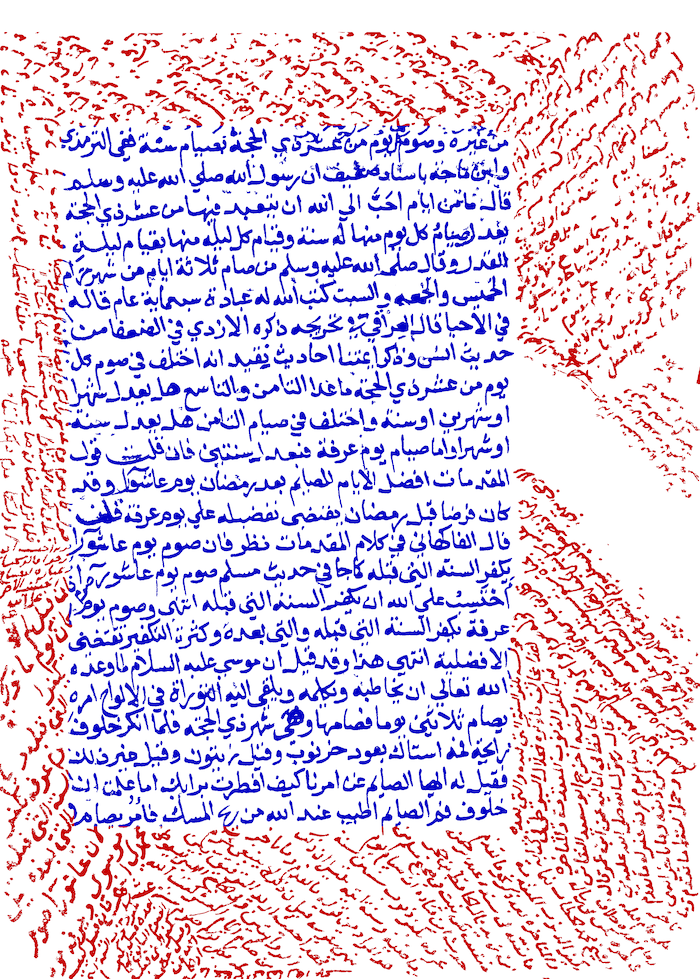}\\
    \end{tabular}
    \caption{Example runs from the test set. First row shows the input image from the test set. Second row, shows visualisation of the feature map. Third, shows mask of the main-text extracted from the feature map. Forth, shows the segmentation result of the method. And the last row shows the groundtruth from the dataset.}
    \label{fig:vis_resutls}
\end{figure}

\figurename~\ref{fig:vis_resutls} shows an example runs of our method. The second row shows a visualization of the extracted feature map using the Siamese network's CNN. The feature map is visualized by mapping the first three principal components to the RGB channels of the image. The visualization show that the CNN were able to extract the meaningful features regarding the main and the side text. 

\begin{table}[!h]
\centering
\caption{Comparison of F-measure values. \cite{bukhari2012layout} and \cite{alaasam2019layout}'s results are with supervised learning and post processing, \cite{barakat2018binarization}'s results are with supervised learning and without post processing whereas our results are with unsupervised learning and without post-processing.}
\begin{tabular}{ p{2.5cm}||p{1.5cm}p{1.5cm} }
\hline
Method & Main text & Side text \\
\hline
Bukhari et al. \cite{bukhari2012layout}       & 95.02  & 94.68 \\
Kurar et al. \cite{barakat2018binarization}   & 95.00  & 80.00 \\
Alaasam et al. \cite{alaasam2019layout}       & \textbf{98.59}  & 96.89 \\
Proposed                                      & 98.56  & \textbf{96.97} \\
\hline
\end{tabular}
\label{tab:results}
\end{table}

\section{Conclusion}
\label{sec:conclusion}
This paper presents an unsupervised page segmentation method for hand-written document images. We train a Siamese network to discriminate between patches with different writing attributes. In addition, the network is trained that two neighboring patches are similar. Our method uses one of the CNN branches of the trained Siamese network to extract a feature map from hand-written document images. The main-text region is extracted based on the first and second principal components of the feature map, which is then used to segment the image into main and side text. We have shown that the proposed method is on par with the supervised state of the art page layout analysis of historical manuscripts in terms of performance. 
In future work, we plan to adapt this method for text line segmentation. In addition, we aim to expand on the idea of using established hand-crafted features to train deep learning networks to tackle other document analysis tasks in an unsupervised setting.

\section*{Acknowledgment}

This research was partially supported by The Frankel Center for Computer Science at Ben-Gurion University.

\bibliography{xref.bib}
\bibliographystyle{ieeetr}

\end{document}